\documentclass[10pt,twocolumn,letterpaper]{article}

\usepackage{cvpr}              

\usepackage{graphicx}
\usepackage{amsmath}
\usepackage{amssymb}
\usepackage{booktabs}

\usepackage{nicefrac} 
\usepackage{xcolor}
\usepackage{graphicx}    
\usepackage{xspace}
\usepackage{tabularx}
\usepackage{multirow}
\usepackage{verbatim}
\usepackage{tabularx}
\usepackage{colortbl}
\usepackage{array, makecell} %
\usepackage{xstring}
\usepackage{placeins}

\usepackage{xspace}
\usepackage[export]{adjustbox}

\usepackage{bold-extra}

\makeatletter

\newcommand{\PAR}[1]{\vskip4pt \noindent {\bf #1~}}

\definecolor{better_blue}{HTML}{4285f4}

\newtoggle{comments}
\toggletrue{comments}
\togglefalse{comments}

\iftoggle{comments}{%
\newcommand{\ali}[1]{\textcolor{blue}{\textbf{Ali: }{#1}}}
\newcommand{\alex}[1]{\textcolor[rgb]{0,0.5,0}{\textbf{Alex: }{#1}}}
\newcommand{\jono}[1]{\textcolor{brown}{\textbf{Jono: }{#1}}}
\newcommand{\deva}[1]{\textcolor[rgb]{0,0.5,0.5}{\textbf{Deva: }{#1}}}
\newcommand{\bastian}[1]{\textcolor[rgb]{0.5,0.5,0}{\textbf{Bastian: }{#1}}}
\newcommand{\ext}[1]{\textcolor{orange}{\textbf{Ext. Rev: }{#1}}}
\newcommand{\todo}[1]{\textcolor{red}{\small Todo:\,#1}\PackageWarning{TODO:}{#1!}}

}
{%
\newcommand{\ali}[1]{}
\newcommand{\alex}[1]{}
\newcommand{\jono}[1]{}
\newcommand{\deva}[1]{}
\newcommand{\bastian}[1]{}
\newcommand{\ext}[1]{}
\newcommand{\todo}[1]{}

 \setlength{\floatsep}{10pt plus2pt minus2pt}
 \setlength{\textfloatsep}{10pt plus2pt minus2pt}
 \setlength{\dblfloatsep}{10pt plus2pt minus2pt}
 \setlength{\dbltextfloatsep}{10pt plus2pt minus2pt}
 \addtolength{\abovecaptionskip}{-5pt}

}

\newcolumntype{P}[1]{>{\centering\arraybackslash}p{#1}}

\usepackage{pifont}

\newcommand{\J}{\mathcal{J}}
\newcommand{\F}{\mathcal{F}}
\newcommand{\JnF}{\mathcal{J}\&\mathcal{F}}

\newcommand{\smalltt}[1]{{\small\texttt{#1}}}

\makeatletter
\@namedef{ver@everyshi.sty}{}
\makeatother
\usepackage{csvsimple}
\usepackage{ifthen}

\newcommand{\cm}[1]{\ifthenelse{\equal{y}{#1}}{\checkmark}{\ifthenelse{\equal{n}{#1}}{}{???}}}
\newcommand{\hq}[2]{\IfSubStr{b}{#2}{\textbf{#1}}{#1}}

\newcommand{\tableline}[2]{
    \csvreader[
        column count=37,
        late after line=\\,
        filter=\equal{\key}{#1}
    ]{tables/data.csv}{
         1=\key,
         2=\fff,
         3=\ytvos,
         4=\framerate,
         5=\backbone,
         6=\GDavisVal, 7=\GDavisValH,
         8=\JDavisVal, 9=\JDavisValH,
        10=\FDavisVal, 11=\FDavisValH,
        12=\GDavisTest, 13=\GDavisTestH,
        14=\JDavisTest, 15=\JDavisTestH,
        16=\FDavisTest, 17=\FDavisTestH,
        18=\GYTVOS, 19=\GYTVOSH,
        20=\JYTseen, 21=\JYTseenH,
        22=\JYTunseen, 23=\JYTunseenH,
        24=\FYTseen, 25=\FYTseenH,
        26=\FYTunseen, 27=\FYTunseenH,
        28=\GYTVOSNew, 29=\GYTVOSNewH,
        30=\JYTNewseen, 31=\JYTNewseenH,
        32=\JYTNewunseen, 33=\JYTNewunseenH,
        34=\FYTNewseen, 35=\FYTNewseenH,
        36=\FYTNewunseen, 37=\FYTNewunseenH
    }{
        & #2 & \cm{\fff} & \hq{\GDavisVal}{\GDavisValH} & \hq{\JDavisVal}{\JDavisValH} & \hq{\FDavisVal}{\FDavisValH} && \hq{\GDavisTest}{\GDavisTestH} & \hq{\JDavisTest}{\JDavisTestH} & \hq{\FDavisTest}{\FDavisTestH} && \hq{\GYTVOSNew}{\GYTVOSNewH} & \hq{\JYTNewunseen}{\JYTNewunseenH} & \hq{\FYTNewunseen}{\FYTNewunseenH} & \hq{\JYTNewseen}{\JYTNewseenH} & \hq{\FYTNewseen}{\FYTNewseenH}
    }
}

\newcommand{\hqg}[2]{\textcolor{gray}{\IfSubStr{b}{#2}{\textbf{#1}}{#1}}}

\newcommand{\tablelineold}[2]{
    \csvreader[
        column count=37,
        late after line=\\,
        filter=\equal{\key}{#1}
    ]{tables/data.csv}{
         1=\key,
         2=\fff,
         3=\ytvos,
         4=\framerate,
         5=\backbone,
         6=\GDavisVal, 7=\GDavisValH,
         8=\JDavisVal, 9=\JDavisValH,
        10=\FDavisVal, 11=\FDavisValH,
        12=\GDavisTest, 13=\GDavisTestH,
        14=\JDavisTest, 15=\JDavisTestH,
        16=\FDavisTest, 17=\FDavisTestH,
        18=\GYTVOS, 19=\GYTVOSH,
        20=\JYTseen, 21=\JYTseenH,
        22=\JYTunseen, 23=\JYTunseenH,
        24=\FYTseen, 25=\FYTseenH,
        26=\FYTunseen, 27=\FYTunseenH,
        28=\GYTVOSNew, 29=\GYTVOSNewH,
        30=\JYTNewseen, 31=\JYTNewseenH,
        32=\JYTNewunseen, 33=\JYTNewunseenH,
        34=\FYTNewseen, 35=\FYTNewseenH,
        36=\FYTNewunseen, 37=\FYTNewunseenH
    }{
        & #2 & \cm{\fff} & \hq{\GDavisVal}{\GDavisValH} & \hq{\JDavisVal}{\JDavisValH} & \hq{\FDavisVal}{\FDavisValH} && \hq{\GDavisTest}{\GDavisTestH} & \hq{\JDavisTest}{\JDavisTestH} & \hq{\FDavisTest}{\FDavisTestH} && \hqg{\GYTVOS}{\GYTVOSH} & \hqg{\JYTunseen}{\JYTunseenH} & \hqg{\FYTunseen}{\FYTunseenH} & \hqg{\JYTseen}{\JYTseenH} & \hqg{\FYTseen}{\FYTseenH}
    }
}

\usepackage[pagebackref,breaklinks,colorlinks]{hyperref}

\usepackage[capitalize]{cleveref}
\crefname{section}{Sec.}{Secs.}
\Crefname{section}{Section}{Sections}
\Crefname{table}{Table}{Tables}
\crefname{table}{Tab.}{Tabs.}

\def\cvprPaperID{1} 
\def\confName{CVPR T4V Workshop}
\def\confYear{2022}

\let\OLDthebibliography\thebibliography
\renewcommand\thebibliography[1]{
  \OLDthebibliography{#1}
  \setlength{\parskip}{0pt}
  \setlength{\itemsep}{0pt plus 0.3ex}
}

\begin{document}

\makeatletter
\DeclareRobustCommand\onedot{\futurelet\@let@token\@onedot}
\def\@onedot{\ifx\@let@token.\else.\null\fi\xspace}

\newcolumntype{Y}{>{\centering\arraybackslash}X}
\newcolumntype{Z}{>{\raggedleft\arraybackslash}X}

\newcommand{\ourMethodName}{HODOR\xspace}


\title{Differentiable Soft-Masked Attention}

\author{
Ali Athar$^1$ %
\quad
Jonathon Luiten$^{1,2}$
\quad
Alexander Hermans$^1$%
\quad
Deva Ramanan$^2$%
\quad
Bastian Leibe$^1$\\[5pt]
$^1$RWTH Aachen University, Germany \quad $^2$Carnegie Mellon University, USA \\[5pt]
{\tt\small \{athar,luiten,hermans,leibe\}@vision.rwth-aachen.de}
\quad {\tt\small deva@cs.cmu.edu} \\
}

\maketitle

\begin{abstract}

Transformers have become prevalent in computer vision due to their performance and flexibility in modelling complex operations. 
Of particular significance is the `cross-attention' operation, which allows a vector representation (e.g. of an object in an image) to be learned by `attending' to an arbitrarily sized set of input features. 
Recently, `Masked Attention' was proposed~\cite{Cheng21ArxivMask2Former,Cheng21ArxivMask2Former_video} in which a given object representation only attends to those image pixel features for which the segmentation mask of that object is active. This specialization of attention proved beneficial for various image and video segmentation tasks.
In this paper, we propose another specialization of attention which enables attending over `soft-masks' (those with continuous mask probabilities instead of binary values), and is also differentiable through these mask probabilities, thus allowing the mask used for attention to be learned within the network without requiring direct loss supervision.
This can be useful for several applications.
Specifically, we employ our `Differentiable Soft-Masked Attention' for the task of Weakly-Supervised Video Object Segmentation (VOS), where we develop a transformer-based network for VOS which only requires a single annotated image frame for training, but can also benefit from cycle consistency training on a video with just one annotated frame. Although there is no loss for masks in unlabeled frames, the network is still able to segment objects in those frames due to our novel attention formulation.
Code: \url{https://github.com/Ali2500/HODOR/blob/main/hodor/modelling/encoder/soft_masked_attention.py}

\end{abstract}

\section{Introduction}
\label{sec:intro}

\begin{figure}[t]
  \centering
  \includegraphics[width=\linewidth]{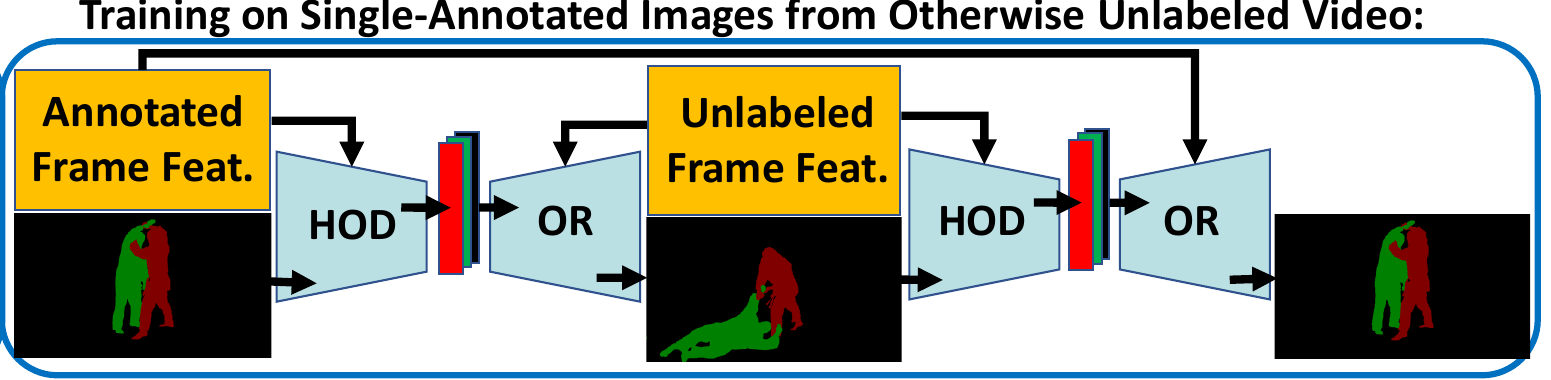}
  \caption{With \textit{Differentiable Soft-Masked Attention}, VOS training can take advantage of unlabeled frames using cycle-consistency. HOD: High-level Object Descriptor Encoder. OR: Object Re-segmentation Decoder.}
  \label{fig:strategies}
\end{figure}

Transformers are revolutionizing computer vision not only due to better performance, but also because they are more flexible. This flexibility enables new classes of applications and modules for processing images, video and 3D representations. 
Transformers allow vector representations to be learned by attending to arbitrarily sized inputs with cross-attention. Recently, \textit{Masked Attention}~\cite{Cheng21ArxivMask2Former, Cheng21ArxivMask2Former_video} was proposed, where the attention operation only attends to features covered by a segmentation mask (usually over pixels of an image). This is beneficial since it accelerates training convergence, enables the network to focus on different segments of the image, and allows these different segments to have arbitrary shapes/sizes e.g. segmentation mask of an object in an image.

However, Masked Attention has two limitations: (1) it requires direct supervision for what the mask should be since the masking operation is not differentiable, and (2) the masks have to be binary. In this paper, we propose a novel formulation for masked attention, which we term \textit{Differentiable Soft-Masked Attention}, which is differentiable and enables attending to mask probabilities instead of just binary mask values. 

\begin{figure*}[t]
  \centering
\includegraphics[width=\linewidth]{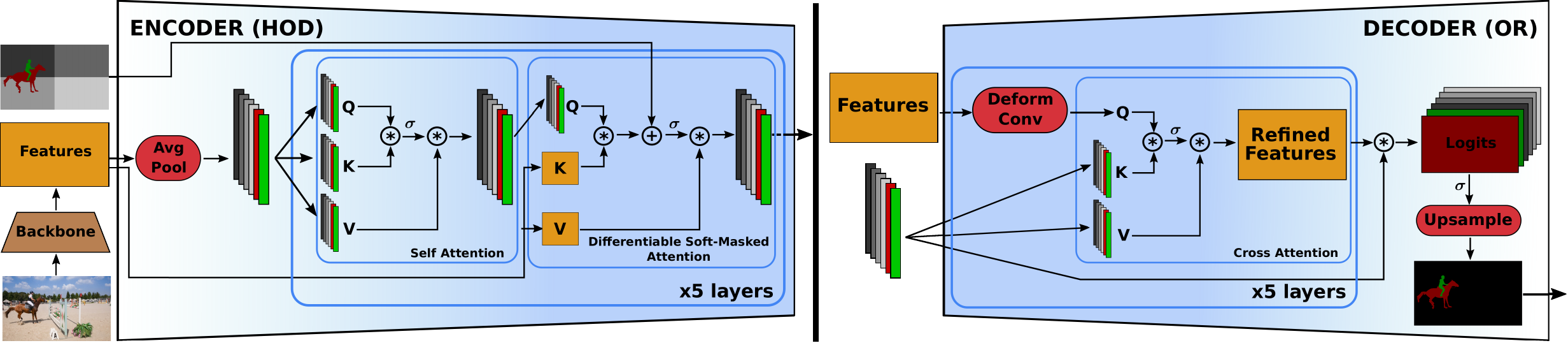}
  \caption{\textbf{The \ourMethodName Architecture} consists of a backbone, the HOD encoder, and the OR decoder. Q, K, and V refer to Queries, Keys and Values, respectively. The encoder jointly encodes all objects and background cells (here $2\times 2$) to descriptors, which are then decoded to masks by the decoder. Some steps are simplified  (the final upsampling) or omitted (fc layers, skip connections). See \cite{HODOR} for details.}
  \label{fig:details}
\end{figure*}

We show through the application of weakly supervised video object segmentation that differentiable soft-masked attention can enable new applications in computer vision. Using this novel attention formulation, we are able to train a network for video object segmentation using video with only a single frame labeled. 
The network is trained by being given an object mask for an object in a video frame. It then learns a high-level descriptor for this object by attending to the image features conditioned on the object mask using our attention formulation. This descriptor can then be propagated to another video frame to segment the object in that unlabelled frame.
We train in a setting where we do not need ground-truth masks for this unlabelled frame. Instead, enabled by our differentiable soft-masked attention, we are able to encode a new object descriptor by attending to the predicted mask in the unlabelled frame, and use this to produce a new mask in the original reference frame where we have the the input ground-truth to apply a loss (see Fig.~\ref{fig:strategies}). Even though there was no direct loss applied to the mask in the unlabelled frame, our network is still able to effectively learn to segment and track objects.
We show that if gradients flow through the predicted segmentation mask, which is exactly the case for our differentiable soft-masked attention, it benefits such a cycle consistent training objective.
This approach achieves impressive and competitive results on both the DAVIS~\cite{Pont-Tuset17Arxiv} and YouTube-VOS~\cite{Xu18Arxiv} benchmarks for video object segmentation, even though no video ground-truth is ever used for training.

This extended abstract is a short version of the paper \textit{HODOR: High-level Object Descriptors for Object Re-segmentation in Video Learned from Static Images} \cite{HODOR}. However, in this short version, we focus only on one particular aspect of the HODOR pipeline, namely the differentiable soft-masked attention, and highlight it for the benefit of the T4V (Transformers for Vision) Workshop Community.

The authors believe that differentiable soft-masked attention has a lot of potential within a number of different computer vision applications, and believe that the T4V workshop is the perfect venue for highlighting this valuable idea within the community.

\section{Differentiable Soft-Masked Attention}
\label{sec:method}

In Mask2Former~\cite{Cheng21ArxivMask2Former,Cheng21ArxivMask2Former_video}, the cross-attention operation attends to pixels which belong to the target image segment. However, this operation is not differentiable w.r.t the mask logits, and it does not permit any interaction between the queries and the background image pixels.

\begin{figure}[t]
  \centering
  \setlength\tabcolsep{1pt}
  \renewcommand{\arraystretch}{0.5}
  \begin{tabularx}{\textwidth}{lccc}%
        \rotatebox[origin=l]{90}{$\alpha = 32$} &
       \includegraphics[width=0.31\linewidth]{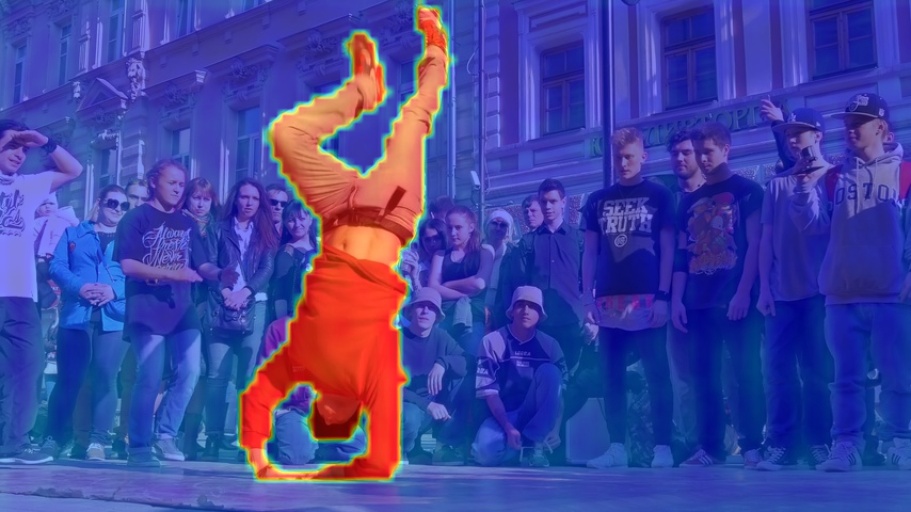}&
       \includegraphics[width=0.31\linewidth]{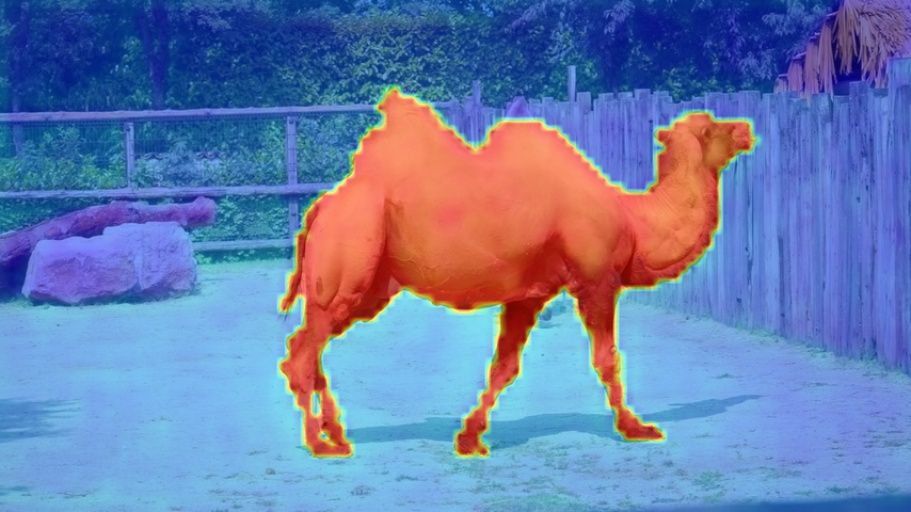}&
       \includegraphics[width=0.31\linewidth]{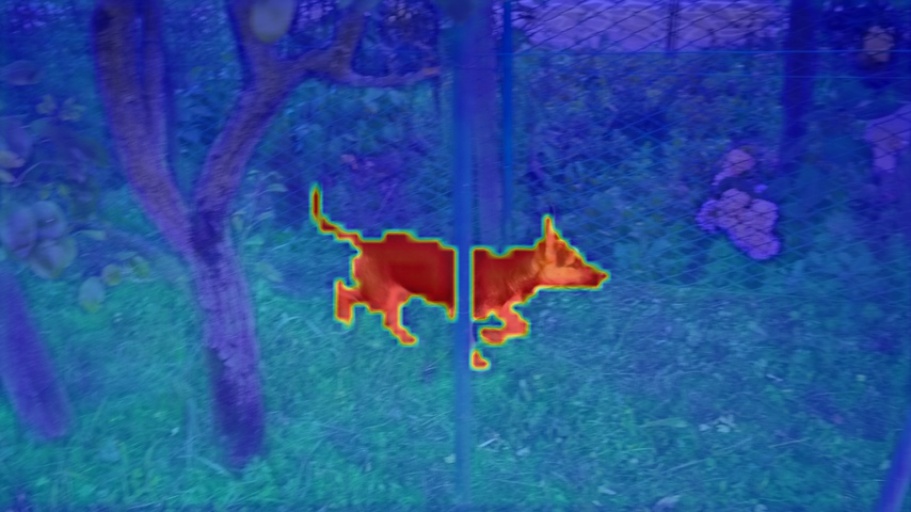}\\
       \rotatebox[origin=l]{90}{$\alpha = 16$} &
       \includegraphics[width=0.31\linewidth]{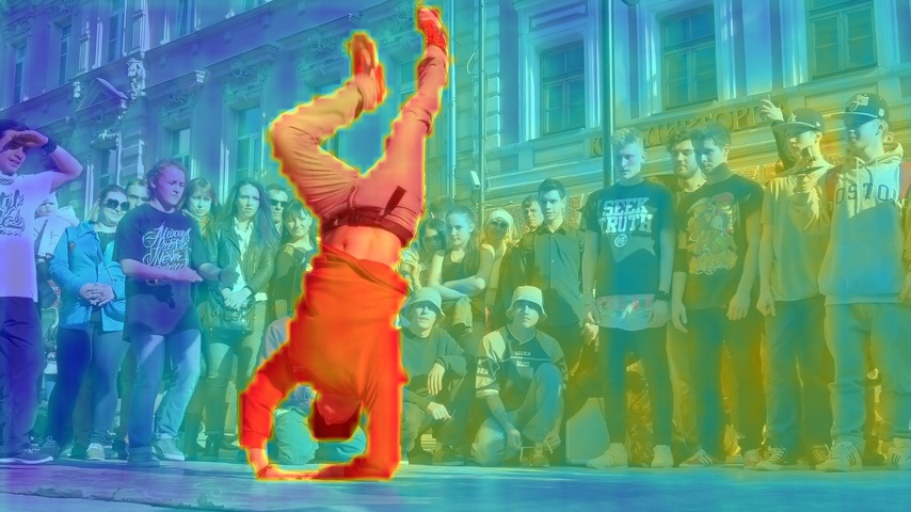}&
       \includegraphics[width=0.31\linewidth]{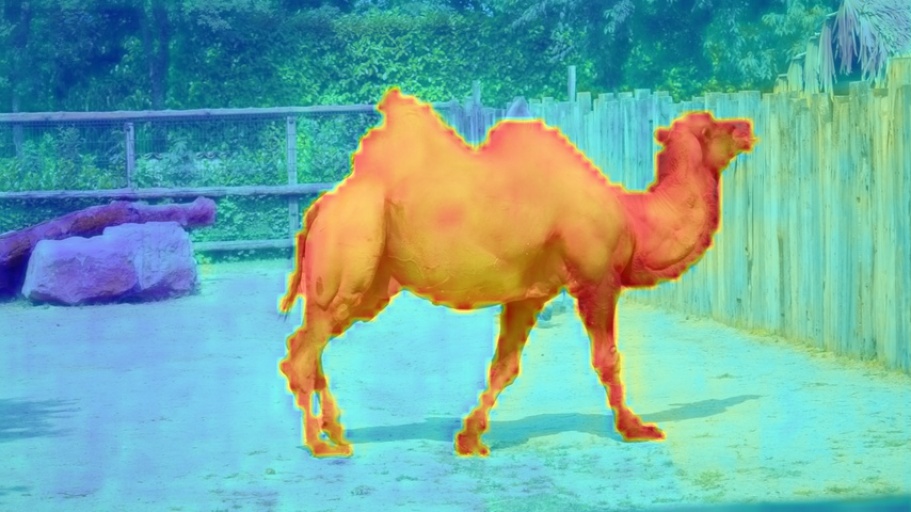}&
       \includegraphics[width=0.31\linewidth]{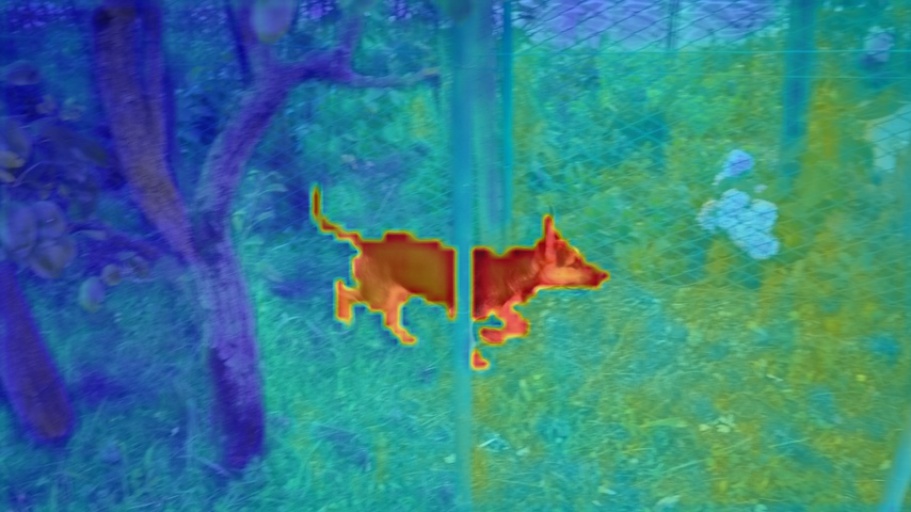}\\
       \rotatebox[origin=l]{90}{$\alpha = 8$} &
       \includegraphics[width=0.31\linewidth]{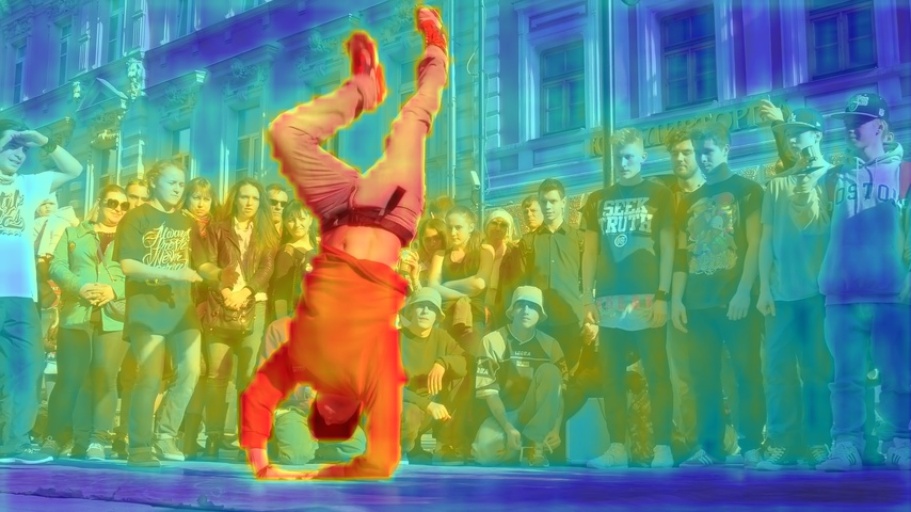}&
       \includegraphics[width=0.31\linewidth]{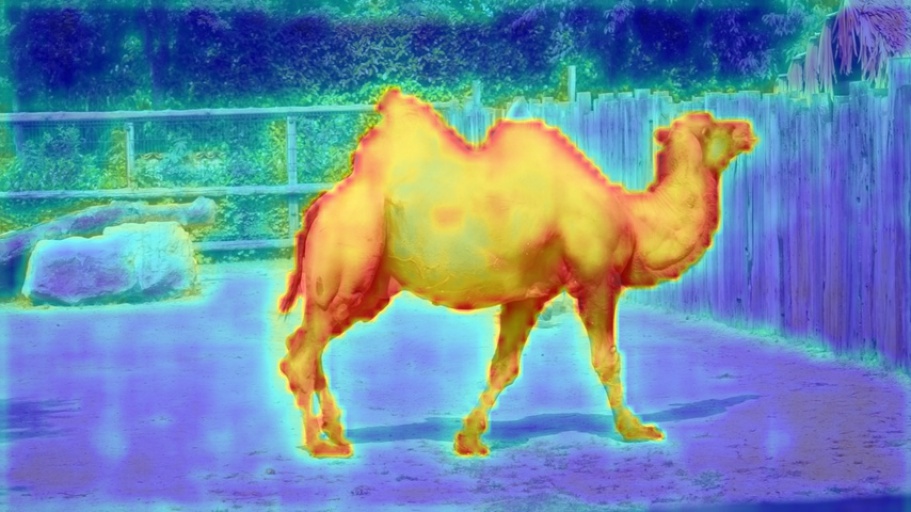}&
       \includegraphics[width=0.31\linewidth]{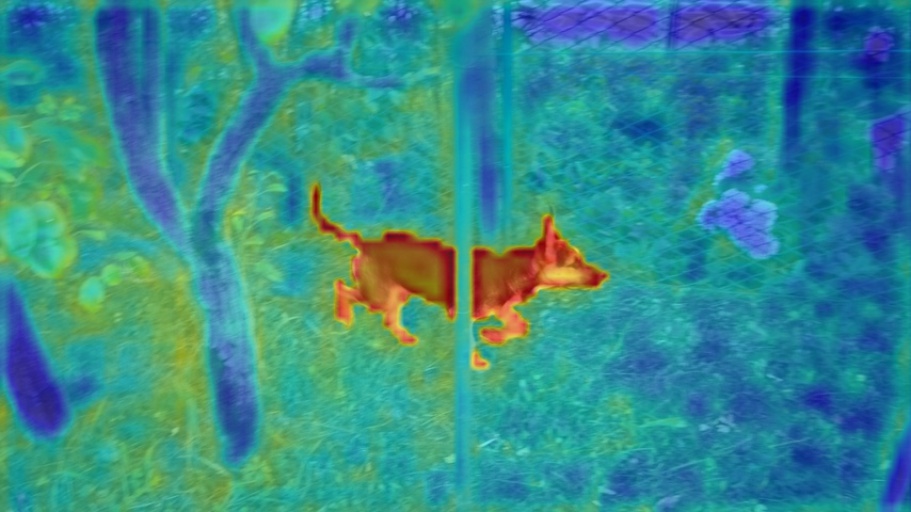}\\
       \rotatebox[origin=l]{90}{$\alpha = 4$} &
       \includegraphics[width=0.31\linewidth]{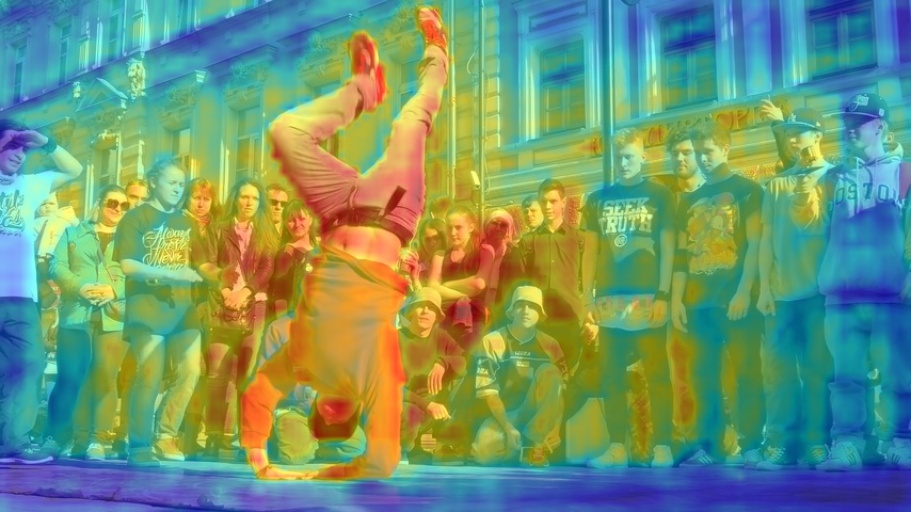}&
       \includegraphics[width=0.31\linewidth]{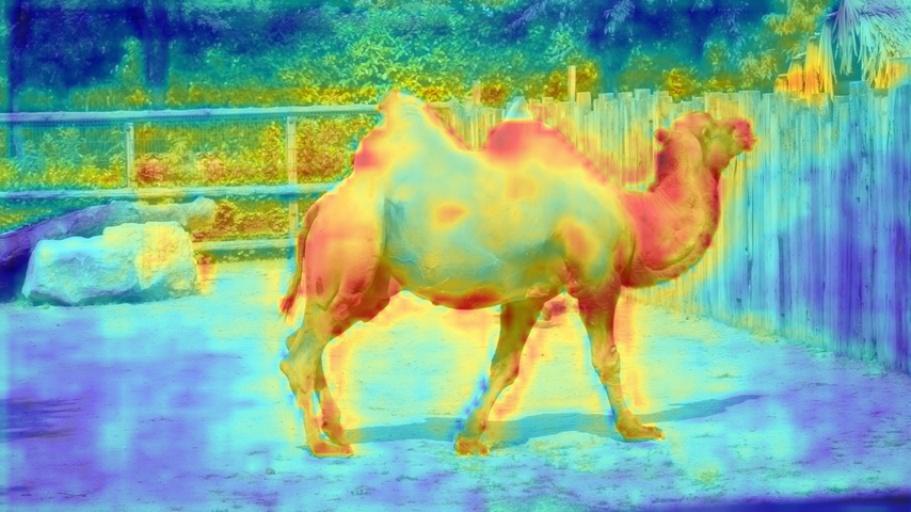}&
       \includegraphics[width=0.31\linewidth]{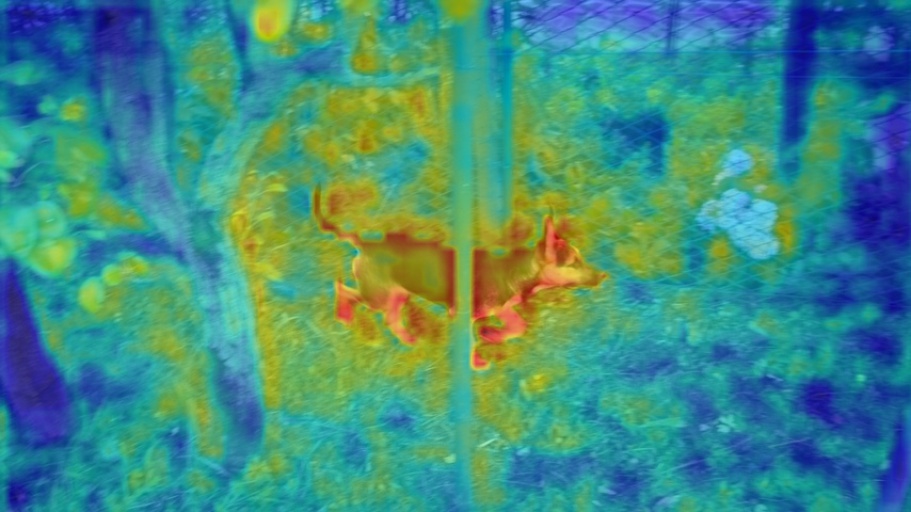}\\
  \end{tabularx}
  \caption{\textbf{Visualization of soft-masked attention weights} Attention weights for the last (5th) encoder transformer layer after masking and before softmax are shown for three sequences from the DAVIS validation set. $\alpha$ is the initialized value of the masking weight (see Eq. 1), this is updated during training.}
  \label{fig:attn_wts}
\end{figure}

We propose a better formulation which is differentiable, 
allows the mask $M$ to be non-binary (\ie have soft values), and also affords the network more flexibility to focus on relevant image features.
Given a mask $M$ with pixel values in the range $[0,1]$, and a learnable, positive scalar $\alpha$, we define the soft masked attention operation as follows:%
\begin{equation}
    \label{eq:encoder_attn_masking}
    \smalltt{softmax} \left( \frac{K^T Q ~{\color{better_blue}+~ \alpha M}} {\sqrt{C}} \right) \cdot V
\end{equation}
This is identical to the standard attention operation proposed by Vaswani~\etal~\cite{Vaswani17NIPS} except for the term `{\color{better_blue}$+ \alpha M$}'. 
In practice, each attention head is assigned a different learnable parameter $\alpha$ which is optimized during training. We initialize $\alpha$ in the 8 heads with ${\alpha=\{ 4,4,8,8,16,16,32,32\}}$. Thus, different attention heads attend to pixel features conditioned on different magnitudes of masking. This enables the network to 
focus on features from the relevant image segment, but also to capture scene information from other parts of the image if this is beneficial for the training objective. This is inspired by Press~\etal~\cite{Press21Arxiv} who used additive offsets in temporal attention in NLP. Fig.~\ref{fig:attn_wts} shows soft-masked attention weights for our trained network.

Note that this formulation contains an inherent information bottleneck which does not allow the input mask's shape or location from directly `leaking' into the output. Specifically, in Eq.~\ref{eq:encoder_attn_masking}, the mask $M$ can only influence the $\smalltt{softmax}(\cdot)$ term, \ie the weights with which the Values ($V$) are summed, but $M$ cannot directly be copied into the attention operation output, which might be detrimental for some applications.

\subsection{Network Architecture}

To demonstrate the benefit of our formulation, we build a Weakly-Supervised Video Segmentation architecture which can take advantage of the properties of our differentiable soft-masked attention. We name this network architecture HODOR: \textit{\underline{H}igh-level \underline{O}bject \underline{D}escriptors for \underline{O}bject \underline{R}e-segmentation}. 

Fig.~\ref{fig:details} illustrates the HODOR network architecture. Here we briefly discuss the workflow, and how differentiable soft-masked attention is applied. For further details, we refer to the full paper~\cite{HODOR}.
%
As shown, HODOR has an encoder-decoder architecture. Given image features from a backbone network and a set of masks for the objects in that image, the encoder learns a concise $C$-dimensional descriptor for each of the objects. The decoder on the other hand performs the dual function: given backbone image features and a set of object descriptors (not necessarily from the same image), it \textit{re-segments} the objects in that image by producing a segmentation mask for each object. 

We employ our soft-masked attention in the encoder where we apply successive layers of cross-attention between the descriptors (queries) and image features (keys/values). Attention masking enables us to effectively condition each descriptor on its corresponding object mask. Moreover, since we can use soft masks, and also back-propagate gradients through them, our formulation enables the encoder-decoder architecture to be applied sequentially across time, \ie given a video sequence where we have mask annotations for objects in the first image frame $I_1$, we encode objects into descriptors, and then decode them into the second frame $I_2$. We can then encode fresh object descriptors using these predicted second frame masks. This process can be repeated throughout the video length. 

\subsection{Training}
\label{subsec:training}

Our attention formulation enables \ourMethodName to be trained using different types of data.

\PAR{Single Images.}
For the basic setting, we only need a static image dataset with annotated object masks. While it is possible to encode/decode the object masks from/into the same image, in practice we train HODOR to propagate object masks through fake `video' sequences generated from single images by applying random affine transformations.

\PAR{Video Cyclic Consistency.}
Our soft-masked attention formulation enables the sequential propagation of object masks over a video to be end-to-end differentiable, \ie even if we only supervise the masks predicted for the last frame of a given clip, the error will be backpropagated over the entire temporal sequence to the first frame. This allows \ourMethodName to also be trained on unlabeled frames from videos with arbitrarily sparse and temporally inconsistent object ID annotations. Given a training clip with $T$ frames where only frame $t=1$ is annotated, we can propagate the given object masks from $t:1 \xrightarrow{} T$, and then further propagate them in reverse temporal order from $t:T \xrightarrow{} 1$. We can then use the principle of cyclic consistency~\cite{Jabri20NeurIPS,Wang19CVPR} for supervision by supervising the predicted masks for $t=1$ to be identical to the input masks.

\begin{table*}[t]
    \centering
        \caption{Quantitative results on the DAVIS and YouTube-VOS datasets.
        For YouTube-VOS we focus on the 2019 validation set, but substitute 2018 validation set results when only those are available (slightly higher, highlighted in 
        \textcolor{gray}{grey}). As is common, we evaluate unseen (us) and seen (s) object classes separately for Youtube-VOS, OL: Online Fine-tuning, $^{\ast}$: retrained by us.}
    \label{tab:main_results}
    
\small
    
\begin{tabularx}{\linewidth}{p{0.3cm}lcYYYcYYYcYYYYY}
\toprule
&&& \multicolumn{3}{c}{DAVIS val 17} && \multicolumn{3}{c}{DAVIS test-dev 17} && \multicolumn{5}{c}{YouTube-VOS \textcolor{gray}{val 18} /val 19}\\
\cmidrule{4-6} \cmidrule{8-10} \cmidrule{12-16}
&  & OL & $\mathcal{J} \& \mathcal{F}$ & $\mathcal{J}$ & $\mathcal{F}$  && $\mathcal{J} \& \mathcal{F}$ & $\mathcal{J}$ & $\mathcal{F}$  && $\mathcal{J} \& \mathcal{F}$ & $\mathcal{J}_{us}$ & $\mathcal{F}_{us}$ & $\mathcal{J}_{s}$ & $\mathcal{F}_{s}$
\\
\midrule

%

%
\tableline{OSVOS}{OSVOS \cite{Caelles17CVPR}}
\tableline{OnAVOS}{OnAVOS \cite{Voigtlaender17BMVC}}
\tableline{OSVOS_S}{OSVOS\textsuperscript{S} \cite{Maninis18TPAMI}}
\tablelineold{STM_im_only}{STM \cite{Oh19ICCV}}
\tableline{DMN_AOA_COCO}{DMN+AOA (COCO) \cite{Liang21ICCV}}
\tableline{KMN_im_only}{KMN \cite{Seong20ECCV}}
\tableline{STCN_im_only}{STCN \cite{Cheng21NeurIPS}}
\tableline{Ours_COCO_aug}{\textcolor{better_blue}{\ourMethodName (Ours)}}
\tableline{Ours_COCO_CC}{\textcolor{better_blue}{\ourMethodName (Ours with CC)}}





\bottomrule

\end{tabularx}
\end{table*}

\section{Experiments}
\label{sec:experiments}

\subsection{Ablations}
\label{subsec:ablations}

We perform ablations to justify the efficacy of our proposed soft-masked attention. These are performed on the DAVIS'17 validation set and the results are reported in Table~\ref{tab:ablations}. 
The evaluation metric is ($\JnF$), which is the mean of the $\J$ score (Jaccard Index) and the $\F$ score ($\text{F}_1$-score). 
We refer to the full paper~\cite{HODOR} for further experiments \cite{HODOR}.

For the first set of experiments, we train on the `Single Images' explained in Sec.~\ref{subsec:training}. When no masking is applied in the cross-attention operation in the encoder, the only cue for the object descriptors to specialize to their respective targets is their initialization (average pooling over the target pixel features, see Fig.~\ref{fig:details}). This setting results in 74.4 $\JnF$. We then apply hard attention masking by thresholding the masks at 0.5 and setting the $\text{K}^T\text{Q}$ matrix entries inside the attention operation to $-\infty$ for pixels where the mask is zero. In other words, this is the same attention masking that is used by Mask2Former~\cite{Cheng21ArxivMask2Former,Cheng21ArxivMask2Former_video}. This strategy yields a similar $\JnF$ of 74.5.
By contrast, with our proposed soft attention masking mechanism, the $\JnF$ improves to 77.5, which is 3\% higher in absolute terms compared to the first two settings.
Finally, we repeat these three experiments for the `Video Cyclic Consistency' training. Here it is worth noting that having no masking in the attention (79.2 $\JnF$) is better than applying hard masking (78.4 $\JnF$).
The performance increase from using our learned soft attention masking in both cases shows that it helps the encoder to better condition the descriptors on the given object masks.



\begin{table}[t]
\centering%
\caption{Several ablation results on the DAVIS 2017 validation set.}%
\label{tab:ablations}%
\small%
  \setlength\extrarowheight{-1pt}
\begin{tabularx}{\linewidth}{llY}
\toprule 
Training Data  & Attention Type & $\mathcal{J\&F}$ \tabularnewline
\midrule 

COCO  & No Masking & 74.4 \tabularnewline
COCO  & Hard Masking & 74.5 \tabularnewline
COCO  & Ours & 77.5 \tabularnewline
\midrule[0.25pt] 
YTVOS+Davis & No Masking & 79.1 \tabularnewline
YTVOS+Davis & Hard Masking & 78.4 \tabularnewline
YTVOS+Davis & Ours & 80.6 \tabularnewline
\bottomrule
\end{tabularx}
\end{table}

\newcommand{\scoreCocoDavisVal}{77.5\xspace}

\subsection{Benchmark Results}
\label{subsec:main_results}

We evaluate \ourMethodName for Video Object Segmentation on the DAVIS'17~\cite{Pont-Tuset17Arxiv} and YouTube-VOS 2019~\cite{Xu18Arxiv} benchmarks where the task is to segment and track an arbitrary number of objects in each video. During inference, the ground truth mask for each object is provided for the first frame in which an object appears.

We compare \ourMethodName to other methods which only use labeled images for training (\ie no annotated video data), and report the results in Table~\ref{tab:main_results}. For more detailed results with methods that use other types of training data, we refer to the full paper~\cite{HODOR} 
Results for \ourMethodName are given for both training setting explained in Sec.~\ref{subsec:training}. We use the same model checkpoint for all three benchmarks.

\ourMethodName trained on COCO achieves 77.5 $\JnF$ on DAVIS'17, outperforming all existing methods. This includes earlier methods~\cite{perazzi2017learning,Caelles17CVPR,Voigtlaender17BMVC,Maninis18TPAMI} that perform online fine-tuning (best score: 68.0 $\JnF$ from $\text{OSVOS}^\text{S}$~\cite{Maninis18TPAMI}), and also current state-of-the-art methods which pre-train on similar augmented image sequences. The best performing method among these is STCN (75.8 $\JnF$) which is 1.7 $\JnF$ lower than our 77.5. 
After fine-tuning on cyclic consistency using only the middle-most annotated frame from each video in the YouTube-VOS and DAVIS training set. This improves the $\JnF$ by 3.1, 1.0, and 0.7 points on the DAVIS validation, DAVIS test, and YouTube-VOS validation sets, respectively.

\section{Conclusion}


We proposed a novel mechanism for differentiable soft attention masking which is used in our network architecture for learning weakly supervised video object segmentation.
Our formulation conditions the cross-attention operation on the soft-valued segmentation mask for a given object, in a way that is differentiable and allows the segmentation mask to be updated via back-propagation. This enables new applications such as learning video segmentation from video with only one frame annotated, because the segmentation mask can be learnt in unannotated frames. Furthermore, this enables the network to focus on both the target object, and also attend to the background features if this is beneficial for optimizing the training objective.
We performed ablations to show that the intuitive flexibility of our approach translates to improved performance on real-world data. Furthermore, we applied our network to three challenging datasets and showed that it out-performs existing methods that only use labeled images for training.

\FloatBarrier

\scriptsize{
\bibliographystyle{abbrv}
\bibliography{abbrev_short,references}
}

\end{document}